\documentclass[sigconf]{acmart}

\usepackage{booktabs} 

\setcopyright{rightsretained}

\usepackage[nolist]{acronym}
\usepackage{enumitem}
\usepackage{xcolor}
\usepackage[normalem]{ulem}
\usepackage{multirow}
\graphicspath{{./images/}}
\DeclareGraphicsExtensions{.pdf,.png} 
\usepackage{colortbl}

\acmDOI{10.1145/3605098.3635887}

\acmISBN{979-8-4007-0243-3/24/04}

\acmConference[SAC '24]{The 39th ACM/SIGAPP Symposium on Applied Computing}{April 8--12, 2024}{Avila, Spain}
\acmYear{2024}
\copyrightyear{2024}

\acmArticle{4}
\acmPrice{15.00}

\acmBooktitle{The 39th ACM/SIGAPP Symposium on Applied Computing (SAC '24), April 8--12, 2024, Avila, Spain}

\begin{document}
\title{Trustful \emph{Coopetitive} Infrastructures for the New Space Exploration Era}
  
\renewcommand{\shorttitle}{\emph{Coopetitive} Infrastructures for Space Exploration}

\author{Lo{\"i}ck Chovet}
\authornote{Both authors contributed equally to this research}
\orcid{0000-0003-4025-9095}
\affiliation{
  \institution{University of Luxembourg}
  \department{SpaceR Research Group, SnT}
  \streetaddress{29, Avenue J.F Kennedy}
  \city{Kirchberg}
  \postcode{L-1855}
  \country{Luxembourg}}
\email{loick.chovet@uni.lu}

\author{Renan Lima Baima}
\orcid{0000-0002-6281-8153}
\authornotemark[1]
\author{Eduard Hartwich}
\orcid{0000-0001-7625-2085}
\affiliation{
  \institution{University of Luxembourg}
  \department{FINATRAX Research Group, SnT}
  \streetaddress{29, Avenue J.F Kennedy}
  \city{Kirchberg}
  \postcode{L-1855}
  \country{Luxembourg}
  }
  \email{renan.limabaima@uni.lu}
  \email{eduard.hartwich@uni.lu}

\author{Abhishek Bera}
\orcid{0000-0002-0196-5969}
\affiliation{
  \institution{University of Luxembourg}
  \department{SpaceR Research Group, SnT}
  \streetaddress{29, Avenue J.F Kennedy}
  \city{Kirchberg}
  \postcode{L-1855}
  \country{Luxembourg}}
\email{abhishek.bera@uni.lu}

\author{Johannes Sedlmeir}
\orcid{0000-0003-2631-8749}
\affiliation{
  \institution{University of Luxembourg}
  \department{FINATRAX Research Group, SnT}
  \streetaddress{29, Avenue J.F Kennedy}
  \city{Kirchberg}
  \postcode{L-1855}
  \country{Luxembourg}}
\email{johannes.sedlmeir@uni.lu}

\author{Gilbert Fridgen}
\orcid{0000-0001-7037-4807}
\affiliation{
  \institution{University of Luxembourg}
  \department{FINATRAX Research Group, SnT}
  \streetaddress{29, Avenue J.F Kennedy}
  \city{Kirchberg}
  \postcode{L-1855}
  \country{Luxembourg}}
\email{gilbert.fridgen@uni.lu}

\author{Miguel Angel Olivares Mendez}
\orcid{0000-0001-8824-3231}
\affiliation{
  \institution{University of Luxembourg}
  \department{SpaceR Research Group, SnT}
  \streetaddress{29, Avenue J.F Kennedy}
  \city{Kirchberg}
  \postcode{L-1855}
  \country{Luxembourg}}
\email{miguel.olivaresmendez@uni.lu}

\renewcommand{\shortauthors}{Chovet and Baima et al.}

\begin{abstract}
  In the new space economy, space agencies, large enterprises, and start-ups aim to launch space \ac{MRS} for various \ac{ISRU} purposes, such as mapping, soil evaluation, and utility provisioning. However, these stakeholders' competing economic interests may hinder effective collaboration on a centralized digital platform. To address this issue, neutral and transparent infrastructures could facilitate coordination and value exchange among heterogeneous space \ac{MRS}. While related work has expressed legitimate concerns about the technical challenges associated with blockchain use in space, we argue that weighing its potential economic benefits against its drawbacks is necessary. This paper presents a novel architectural framework and a comprehensive set of requirements for integrating blockchain technology in MRS, aiming to enhance coordination and data integrity in space exploration missions. We explored \ac{DLT} to design a non-proprietary architecture for heterogeneous \ac{MRS} and validated the prototype in a simulated lunar environment. The analyses of our implementation suggest global \ac{ISRU} efficiency improvements for map exploration, compared to a corresponding group of individually acting robots, and that fostering a \emph{coopetitive} environment may provide additional revenue opportunities for stakeholders.
\end{abstract}




\begin{acronym}
\acro{DLT}[DLT]{distributed ledger technology}
\acro{ISRU}[ISRU]{in-situ resource utilization}
\acro{MRS}[MRS]{multi-robot systems}
\acro{ESA}[ESA]{european space agency}
\acro{ESRIC}[ESRIC]{european space resources innovation centre}
\acro{NFT}[NFT]{non-fungible tokens}
\acro{IPFS}[IPFS]{InterPlanetary File System}
\acro{DSR}[DSR]{design science research}
\acro{DSRM}[DSRM]{design science research method}
\acro{EL3}[EL3]{european cooperative large logistic lander}
\end{acronym}


\received{29 September 2023}
\received[revised]{10 December 2023}
\received[accepted]{05 January 2024}

\maketitle

\section{Introduction}
\label{sec:introduction}

The space industry has experienced substantial growth in recent years, pushed by various government policies, international agreements~\cite{leonMiningMeaningExamination2018}, and private investments, accounting for 80\,\% of the sector's revenue over the last 16~years~\cite{connGlobalSpaceEconomy2021}. Major space agencies, such as~\citet{nasaSpaceTechnologyRoadmaps2016}, have emphasized the emergence of long-term human presence on space constellations and \ac{MRS} lunar missions for destination reconnaissance, resource prospecting, and mapping~\cite{crawfordLunarResourcesReview2015, schusterARCHESSpaceanalogueDemonstration2020, armScientificExplorationChallenging2023, lorenzHowFarFar2020}, also known as \ac{ISRU}. Combining this tendency with recent regulatory framework changes — shifting from a strict antitrust approach to a more consumer-focused free market~\cite{orlovaPresentFutureSpace2020} — highlights the potential for further expansion and advancement toward a market-based space exploration approach. As the space industry moves towards decentralization and a collaborative approach, there is a growing need to explore innovative solutions that enable efficient coordination among multiple robots~\cite{quintonMarketApproachesMultiRobot2023a}.

\Ac{MRS} involve groups of robots working together or supporting each other to accomplish specific tasks~\cite{parkerMultipleMobileRobot2008}. These systems can involve \emph{homogeneous} or \emph{heterogeneous} groups of robots and are traditionally categorized by their level of goal similarity, awareness of each other, and interaction as \emph{collective}, \emph{cooperative}, or \emph{collaborative}~\cite{parkerMultipleMobileRobot2008}. The emergence of \emph{coopetitive} systems, where competing agents simultaneously choose to cooperate owing to economic incentives, further expands the possibilities of \ac{MRS}~\cite{zhangGroupsymmetricConsensusNonholonomic2022} in \ac{ISRU} driven explicitly by economic incentives and shared objectives. Information-sharing protocols and market-based approaches can often improve coordination among \ac{MRS} and robots’ resource utilization, cost-effectiveness, and exploration capabilities~\cite{quintonMarketApproachesMultiRobot2023a}. These systems can effectively mitigate adverse selection caused by information asymmetries — situations where few market participants know more about products (e.g., water or iron positions) or service quality (e.g., mapping data) than others~\cite{akerlofMarketLemonsQuality1970}, and as such, contribute to efficiency in the market~\cite{hartwichMachineEconomies2023} in line with the growing acceptance of this technology in space missions~\cite{european_space_agency_blockchain_2019}.

Coordinating \ac{MRS} in a market-based approach towards \ac{ISRU} presents significant challenges due to the involvement of multiple competing entities and nations, with more than 60~countries involved in space activities~\cite{borowitzStrategicImplicationsProliferation2019}.
Additionally, the legal~\cite{smithArtemisProgramOverview2020} and technical requirements for planetary mobility systems further complicate the coordination of \ac{MRS} in space exploration~\cite{lorenzHowFarFar2020}. This work's technical and economic foundations highlight the need for robust and adaptable \ac{ISRU} systems that disincentivize undesirable behavior~\cite{leonMiningMeaningExamination2018}.

Additionally, stakeholders involved in space missions highly value the achievement of being the first to explore and acquire space and scientific data, so companies should aim to make mission outputs (e.g., videos, images, and audio) broadly accessible (e.g., via the web and media)~\cite{cameronGoalsSpaceExploration2011}. This perspective resonates with the principles of open science, which promotes knowledge sharing, transparency, collaboration, and accessibility in research~\cite{mckiernanHowOpenScience2016}.
The evolving role of space ecosystems in shaping the future of space exploration~\cite{orlovaPresentFutureSpace2020} underscores the need for a decentralized (i.e., non-proprietary), trustworthy, and transparent digital platform. Such a platform can facilitate the seamless exchange of information and value among stakeholders~\cite{hoessBlockchainEffectInterEcosystem2021}, enabling autonomous \ac{MRS} coordination for resource trading. \ac{DLT}-based systems might offer a suitable solution for this specific requirement~\cite{butijnBlockchainsSystematicMultivocal2020b} and, as a consequence, have been considered the foundation of both space \ac{MRS}~\cite{filippiBlockchainOuterSpace2021} and open science platforms~\cite{coelhoIntegratingBlockchainData2021}.

\ac{DLT} offers several advantages for machine cooperation, including in \ac{MRS} settings~\cite{zivicDistributedLedgerTechnologies2019, hartwichMachineEconomies2023}. These platforms can automate tasks such as bidding for resource usage, publicly broadcasting resource acquisition, and facilitating immediate operational cost compensation. With a \ac{DLT}-based digital platform, entities can provide idle robot resources and stack intermediary profitable tasks to automatically compensate operational costs and openly recognize pioneering exploratory participants, thus increasing \ac{ISRU} efficiency and enabling lower-cost space exploration. Despite the advantages, \ac{DLT} in space \ac{MRS} is not without challenges. For instance, space robots face harsh conditions and limited resources, which conflict with the inefficient information processing of blockchains' intensive computation and storage replication~\cite{butijnBlockchainsSystematicMultivocal2020b, guggenberger2022fabric}. Furthermore, despite stakeholders' interest in open science~\cite{cameronGoalsSpaceExploration2011}, replicated information processing must still be aligned with the need to protect the sensitive business information of robots or organizations exposed through transactional (meta-) data~\cite{sedlmeir2022transparency}.

Therefore, tensions exist between the opportunities and challenges of using \ac{DLT} for space \ac{MRS}, requiring closer investigation. This paper aims to contribute to this understanding by designing an architecture for \emph{coopetitive} \ac{MRS} for \ac{ISRU}, focusing on facilitating open science through \ac{DLT}. We drew upon the ESA-ESRIC space resources challenge as a specific use case~\cite{linkESAESRICSpaceResources2021} of mapping exploration and applied \ac{DLT} to coordinate automated cross-organizational economic interactions effectively. Our study investigates the technical feasibility of leveraging \ac{DLT} to enable participation from universities, research institutes, and small companies in low-cost explorative scientific research. Such a decision-making platform must be holistically assessed to determine whether or not it can improve the global efficiency of \ac{ISRU}~\cite{diasMarketBasedMultirobotCoordination2006, orlovaPresentFutureSpace2020}, maintain targeted information symmetry, and compensate space companies to create additional revenue streams for their (idle) robotic resources. Hence, our research question (RQ) is as follows:

\textbf{RQ:} \textit{Can an \ac{ISRU} platform architecture based on \ac{DLT} address the challenges of coordinating \emph{coopetitive} space \ac{MRS} mapping for open science?}

In the evolving landscape of space exploration, which increasingly favors collaborative approaches~\cite{orlovaPresentFutureSpace2020}, our case study, detailed in Section~\ref{sec:scenario}, motivates the need for an architecture that empowers organizations to utilize \ac{MRS} capabilities for efficient \ac{ISRU} data acquisition. Adopting the \ac{DSR} method by Peffers et al.~\cite{peffersDesignScienceResearch2007d} to answer our \textbf{RQ}, we have conceptualized a decentralized (i.e., non-proprietary) \ac{MRS} platform. This platform, elucidated in Section~\ref{sec:solution}, is grounded in a thorough literature review, integrating insights from \ac{DLT}~\cite{butijnBlockchainsSystematicMultivocal2020b}, space \ac{MRS}~\cite{gaoReviewSpaceRobotics2017}, \ac{DLT} in robotics~\cite{aditya2021survey}, \ac{DLT} in distributed control and cooperative robots~\cite{khanBlockchainTechnologyApplications2019}, and \ac{DLT} for the space industry~\cite{ibrahimLiteratureReviewBlockchain2021}. As outlined in Section \ref{sec:solution}, we develop a \ac{DLT}-based \emph{coopetitive} \ac{MRS} that supports space exploration, primarily scientific \ac{ISRU} mapping. These objectives were shaped by an in-depth case study~\cite{eisenhardtTheoryBuildingCases2007a} and a literature review to discern the system's requirements. By leveraging \ac{DLT}, we strive to establish a platform that ensures verifiable exchanges of information and value among diverse, untrusted stakeholders. Our iterative development and evaluation processes, described in Sections~\ref{sec:solution} and~\ref{sec:evaluation}, have been instrumental in refining the platform's design to meet these objectives and address \emph{coopetitive} \ac{ISRU} challenges. The evaluations, conducted in three diverse environments, as detailed in Section~\ref{sec:evaluation}, have been pivotal in assessing the platform's efficacy and adaptability. The limitations and open challenges are discussed in Section~\ref{sec:discussion}. Section~\ref{sec:conclusion} highlights the platform's potential to enhance the efficiency of \ac{ISRU} activities and create new revenue streams for stakeholders. Our findings and supplementary resources will be publicly available on platforms like GitHub and YouTube.
\section{Scenario: Requirements Elicitation}
\label{sec:scenario}

This section delves into our space mapping scenario that utilizes decentralization and \emph{coopetition} to enhance efficiency in mapping coordination among multiple robots. The purpose is to outline the scope and specific use case we will refer to throughout the paper. Section~\ref{sec:discussion} presents further discussions and limitations of our artifact. We employed the decentralized multi-robotic platform REALMS2 \cite{vandermeerREALMSRESILIENTEXPLORATION2023}, it uses three Leo~Rovers\textsuperscript{\textregistered}~\cite{karalekasEUROPACaseStudy2020} and a \mbox{LUVMI-X} robot to identify resources and analyze and map the environment~\cite{garcetLunarVolatilesMobile2019}.

\begin{figure}
	\centering
	\includegraphics[width=\columnwidth]{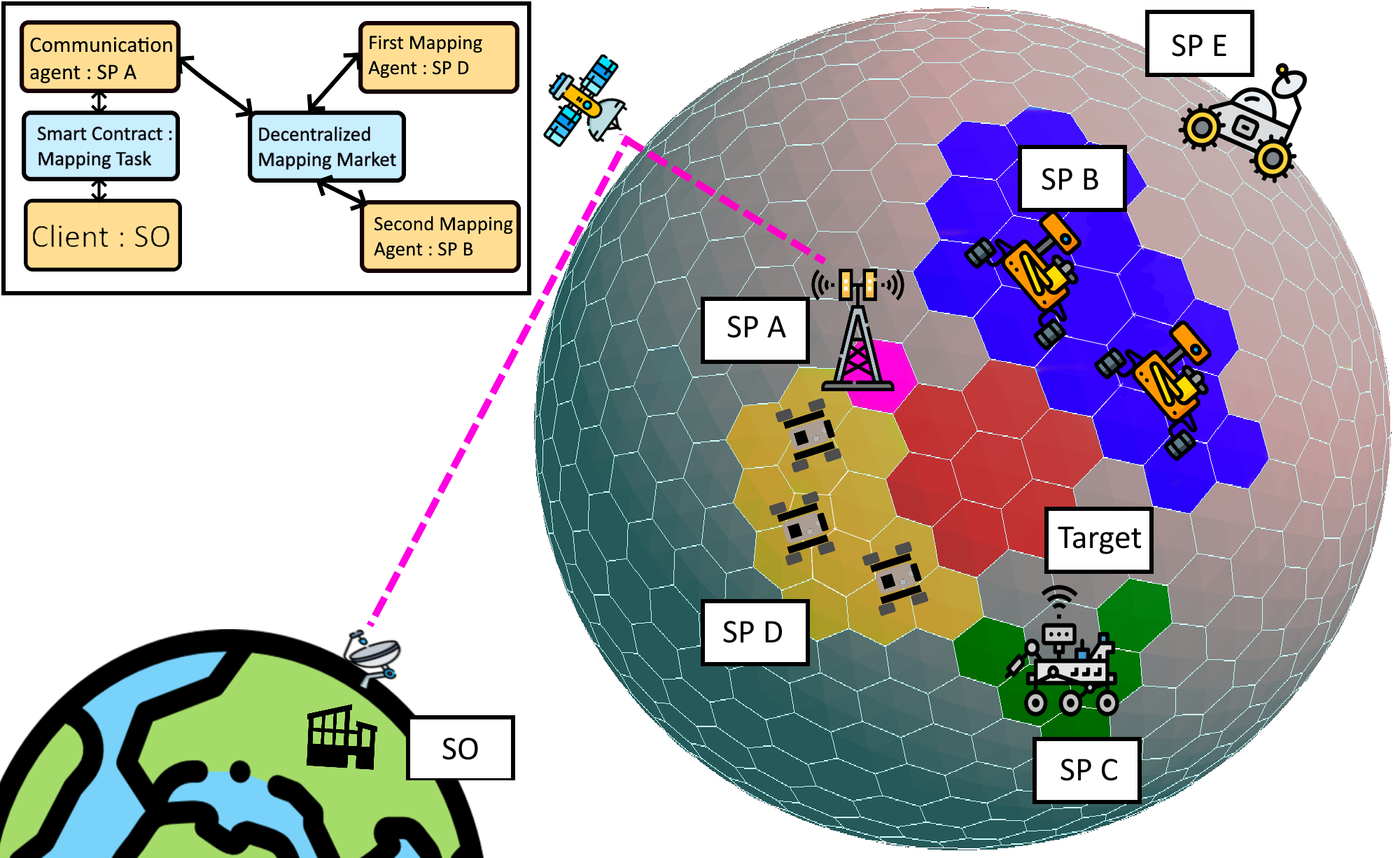}
	\caption{Moon's Goldberg polyhedron diagram, each zone color represents a company's \ac{MRS} operational area~\cite{gooDIANABlockchainLunar2019}.}
    \label{fig:Moonmap}
    \Description{A diagram depicting the astrological constellation arranged in a Goldberg polyhedron format. This surface comprises hexagons and pentagons and serves as a geographical positional of the NFTs. Over the surface, various types of symbolic robots and satellites represent the interacting \ac{MRS}. Each system has its name and color-coded areas (yellow, green, and blue) for easy identification. The target to be mapped is color-coded as red.}
\end{figure}

To illustrate this setting more precisely, we revisit the market-based mapping scenario from~\citet{diasMarketBasedMultirobotCoordination2006} and illustrate the result in Figure~\ref{fig:Moonmap}. This approach, proven effective in patrolling, exploration, and pick-and-delivery~\cite{quintonMarketApproachesMultiRobot2023a}, is grounded in market-based strategies for \ac{MRS}. For the sake of simplification, we are using exemplary amounts in the following scenario. The celestial surface stratification follows the existing Goldberg polyhedron approach~\cite{gooDIANABlockchainLunar2019} and uses the selenographic system to refer to its surface positions. An initiator stationed on Earth, the service orderer~(SO), sends requests to a celestial stationed network of \ac{MRS} service providers~(SP), as listed in Table~\ref{tab:ListSP}. The SO proposes to pay a maximum of \$50 for mapping the target region. SP~D is 5\,m away, while SP~C is 10\,m from the target area. We consider an oversimplified abstraction of each robot's cost function, with a cost of \$2 for each meter they travel. Consequently, SP~D wins the contract by bidding \$10 compared to \$20.%

\begin{table}
\centering
\caption{List of service providers.}
\label{tab:ListSP}
\resizebox{\columnwidth}{!}{
\begin{tabular}{llllp{0.3\columnwidth}} 
\toprule
SP & Color & Robotic Fleet & Price & \multicolumn{1}{c}{Main Focus} \\ 
\midrule
A & Pink & \begin{tabular}[t]{@{}l@{}}Communications satellites,\\ Multiple antennas\end{tabular} & - & \multicolumn{1}{l}{\begin{tabular}[t]{@{}l@{}}Moon-earth\\communications\end{tabular}} \\
B & Blue & Medium size offline fleet of robots & - & Mapping \\
C & Green & Few robots w/ embedded sensors & \$20 & Resource analysis \\
D & Yellow & Large fleet of small robots & \$10 & \multicolumn{1}{l}{\begin{tabular}[t]{@{}l@{}}Fast mapping with\\less precision\end{tabular}} \\
E & None & A single robot & \$200 & Mapping \\
\bottomrule
\end{tabular}
}

\end{table}

The primary need of \ac{ISRU} SOs is to locate and investigate resources from celestial bodies~\cite{cameronGoalsSpaceExploration2011}. This involves identifying areas with water or ice and mapping out the geological features of the planet's terrain. This role could often be triggered and motivated by universities and research institutions that seek to advance scientific knowledge and innovation in space. However, the scarcity of resources available to SOs due to high budget requirements for space-related research hampers advancements in this field. Incomplete mapping data is also an issue owing to the limited coverage of celestial geography~\cite{crawfordLunarResourcesReview2015}. Access to more precise information about celestial resources and terrain would significantly improve scientific outcomes~\cite{lorenzHowFarFar2020}.

The SPs, on the other hand, may represent start-ups~\cite{crawfordLunarResourcesReview2015, schusterARCHESSpaceanalogueDemonstration2020}, large companies~\cite{satellitesSESSatelliteTelecommunications2021}, space agencies~\cite{nasaSpaceTechnologyRoadmaps2016}, and organizations that already had some level of collaboration internally but not necessarily with one another. These entities are willing to work together to maximize their revenues and return on mission investment by offering resources, such as idle robots, shelter time, or energy supply. However, establishing trustworthy partnerships across various knowledge domains~\cite{kannengiesserChallengesCommonSolutions2022}, including international agreements~\cite{smithArtemisProgramOverview2020, orlovaPresentFutureSpace2020} and individual initiatives~\cite{israelSpaceGovernance2020, connGlobalSpaceEconomy2021, schusterARCHESSpaceanalogueDemonstration2020}, poses a challenge. Their struggle with inefficient use of exploration units, methods, and resources often results in wasted opportunities and increased costs~\cite{yan2013survey}. The most significant values of these entities are in guaranteeing mission pioneering and promptly sharing exploratory data~\cite{cameronGoalsSpaceExploration2011}. By following open science principles, they can encourage collaboration and innovation and gain access to valuable space exploration data~\cite{mckiernanHowOpenScience2016}.
We do not claim that these requirements are fully comprehensive or sufficient, yet we consider them necessary.

\section{Foundations}
\label{sec:foundations}

Existing solutions for heterogenous \ac{MRS} exploring the moon are still in their development phase. As an example of current advancements, Corob-X \cite{dettmannCOROBXCooperativeRobot2022} represents a notable initiative in the domain of space robotics. However, it is important to note that Corob-X is a proprietary solution that relies on mutual collaboration and trust among stakeholders. In contrast, our approach offers a non-proprietary, blockchain-based framework, addressing the need for decentralized trust and open collaboration in \ac{MRS}.

\subsection[Market-based coordination for coopetitive MRS and space]{Market-based coordination for \emph{coopetitive} \ac{MRS} and space}

Coordinating multiple robots in market-based systems has been a focal point of extensive research evaluating their quality, trade efficiency, and impact on performance through theoretical analysis and experimental results~\cite{diasMarketBasedMultirobotCoordination2006, quintonMarketApproachesMultiRobot2023a}. Centralized approaches often overlook the complexities of cross-organizational interactions~\cite{canezDevelopingFrameworkMake2000}. In contrast, emerging distributed and decentralized market-based \ac{MRS} offer advantages in optimizing capital distribution, supply and demand matching, and automated resource allocation while reducing risks associated with centralized systems, such as single points of failure or direct peer-to-peer setups~\cite{kapitonovRobonomicsBasedBlockchain2018}.

Auction-based mapping systems, though less complex than traditional solutions, require greater participation and have been improved by integrating contractual models to enhance efficiency in the space industry~\cite{diasMarketBasedMultirobotCoordination2006,belyaevaContractualArrangementsProviders2021}.
Integrating these contracts into digital platforms supported by reliable networks can enhance peer autonomy and participation~\cite{filippiBlockchainOuterSpace2021, ibrahimLiteratureReviewBlockchain2021}. Given the vulnerabilities of centralized systems to censorship and manipulation~\cite{calvaresiTrustedRegistrationNegotiation2018}, combining market-based approaches with blockchain technology offers a promising avenue for enhancing \ac{MRS} coordination.

Decentralized approaches using \ac{DLT} have emerged, with some employing blockchain-based auction systems to determine resource usage in competitive scenarios and ensure trustworthy services~\cite{kapitonovRobonomicsBasedBlockchain2018}. Such approaches promote the efficiency, robustness, adaptability, fault tolerance, effectiveness, and responsiveness of \ac{MRS}~\cite{kapitonovRobonomicsBasedBlockchain2018, zlotMultirobotExplorationControlled2002, garciaArchitectureDecentralizedCollaborative2018, garciaRoboticsSoftwareEngineering2020}. However, fully distributed systems sometimes produce suboptimal solutions~\cite{diasMarketBasedMultirobotCoordination2006, quintonMarketApproachesMultiRobot2023a}. Thereby, given that the technical requirements for planetary mobility, such as resource and latency constraints~\cite{lorenzHowFarFar2020, guggenberger2022fabric}, add another layer of complexity, these challenges necessitate innovative, real-time coordination solutions in \ac{ISRU} scenarios. While existing research has primarily focused on single-entity settings~\cite{castelloferrerBlockchainNewFramework2019}, there remains a gap in the literature for blockchain and auction-based \ac{MRS} solutions in \emph{coopetitive} \ac{ISRU} scenarios. %

\subsection[Distributed ledger technology for space]{\Acl{DLT} for space}

\ac{DLT} provides decentralized transaction recording and validation across multiple nodes, offering similar properties as common digital platforms but without a central operator~\cite{cataliniSimpleEconomicsBlockchain2020}. Blockchain, a subset of \ac{DLT}, is characterized by its replicated, append-only, hash-linked data structures~\cite{butijnBlockchainsSystematicMultivocal2020b}. The technology allows for various levels of openness regarding participation and access, ranging from permissionless to permissioned and public to private networks~\cite{beck2018governance}.
Blockchain's non-proprietary nature enhances trust among entities and enables programmable value exchanges using fungible and non-fungible tokens~\cite{hartwichProbablySomethingMultilayer2023}.
Smart contracts, often using standards like ERC20, facilitate standardized transactions and programming logic ~\cite{meitingerSmartContracts2017,ansariImplementationEthereumRequest2020}
Using blockchain technology to conduct direct peer-to-peer transactions and facilitate smart contracts can reduce costs and increase productivity in various industries~\cite{afanasyevBlockchainSolutionsMultiAgent2019, zivicDistributedLedgerTechnologies2019, hoessBlockchainEffectInterEcosystem2021}. Our work explores the potential benefits and limitations of implementing \ac{DLT} in the identified gap of \ac{MRS} for \ac{ISRU}. We focus on enhanced data sharing and the automation of services using smart contracts, thereby creating new revenue opportunities, such as offering robots idle time and promoting non-discriminatory participation among stakeholders, ultimately establishing a \emph{coopetitive} \ac{ISRU} economy.

\section{Solution: requirements and architecture}
\label{sec:solution}

We conducted a case study and literature review to collect design requirements for our \emph{coopetitive} \ac{MRS} based on \ac{DLT} and focused on \ac{ISRU}, as described here and in sections~\ref{sec:scenario} and~\ref{sec:foundations}. Although the requirement list is not final, we integrated insights from the previous sections' discussions to create a comprehensive list addressing the unique challenges and complexities of implementing \ac{DLT} in an \ac{MRS} for space. Our system architecture draws inspiration from the works of~\citet{liBlockchainBasedCrowdsourcingFramework2022}, \citet{ryosukeabeFabchainManagingAuditable2022}, and requirements from~\citet{lorenzHowFarFar2020} and~\citet{cameronGoalsSpaceExploration2011}. It follows the make-or-buy economic framework~\cite{canezDevelopingFrameworkMake2000} while prioritizing transparency principles of open science~\cite{mckiernanHowOpenScience2016} and the cost-efficiency characteristics of \emph{coopetitive} settings~\cite{ghobadiCoopetitiveKnowledgeSharing2011b}. Our technical design enables geographically distributed robots to self-coordinate~\cite{garciaArchitectureDecentralizedCollaborative2018} by using market-based strategies for service provisioning~\cite{quintonMarketApproachesMultiRobot2023a, diasMarketBasedMultirobotCoordination2006, zlotMultirobotExplorationControlled2002}.

\subsection{Requirements analysis}

Our design aims to generate new revenue streams through map selling and granting recognition as pioneer explorers by meeting the following requirements:
\begin{enumerate}[label=\roman*.]
    \item \textbf{Network:} Enabling clients to create job postings and broadcast them through the mesh network, enabling robots to communicate and coordinate tasks trustfully~\cite{lorenzHowFarFar2020, cameronGoalsSpaceExploration2011}.
    \item \textbf{Data sharing transparency:} Transparency about who shares which information at which time -- by monitoring explorers and job execution history -- fosters a \emph{coopetitive} environment with reduced information asymmetry~\cite{ghobadiCoopetitiveKnowledgeSharing2011b, mckiernanHowOpenScience2016}.
    \item \textbf{Robot agnostic system:} Compatibility with any robotic platform following the established norms, with the capacity to scale the system~\cite{coelhoIntegratingBlockchainData2021, quintonMarketApproachesMultiRobot2023a}.
    \item \textbf{Data loss resistance:} Ensuring data remains intact and accessible despite local system failures or network disruptions, safeguarding valuable information~\cite{belyaevaContractualArrangementsProviders2021}, e.g., operational map data.
\end{enumerate}

\subsection{Requirements specification}

To be more precise, the significant functional and non-functional requirements are identified as follows:

\subsubsection{Functional requirements:}
\begin{enumerate}
\item Receive and process robots and SOs' map requests~\cite{zlotMultirobotExplorationControlled2002}.
\item Implement a descending-price auction mechanism for robots to allocate mapping tasks~\cite{zlotMultirobotExplorationControlled2002, quintonMarketApproachesMultiRobot2023a}.
\item Verify the integrity and completeness of metadata associated with each job request~\cite{diasMarketBasedMultirobotCoordination2006}.
\item Execute the payment process for completed tasks~\cite{diasMarketBasedMultirobotCoordination2006}.
\item Provide an interface for managing and monitoring job requests during the mission~\cite{diasMarketBasedMultirobotCoordination2006}.
\item Enable mapping and \ac{ISRU} performance assessment~\cite{garciaArchitectureDecentralizedCollaborative2018, lorenzHowFarFar2020}.
\item Ensure an adaptable robotic network compatible with the different robot types~\cite{diasMarketBasedMultirobotCoordination2006, zlotMultirobotExplorationControlled2002, garciaArchitectureDecentralizedCollaborative2018}.
\item Support the seamless integration with possible existing infrastructures during space missions, such as space decentral~\cite {spaceConsenSysSpace2018}, SpaceChain~\cite{zhengSpaceChainCommunitybasedSpace2018}, and TruSat initiatives~\cite{mousaviniaSpaceDecentralDecentralized2012, garciaArchitectureDecentralizedCollaborative2018}.
\item Ensure compatibility with standard data formats and protocols for efficient data exchange~\cite{garciaArchitectureDecentralizedCollaborative2018}.
\item Promote interoperability of robotic communication protocols to facilitate seamless coordination~\cite{garciaArchitectureDecentralizedCollaborative2018}.
\end{enumerate}

\subsubsection{Non-functional requirements:} 
\begin{enumerate}
\item Conformity: Provide an interoperable economic framework interface for efficient market-based coordination~\cite{zlotMultirobotExplorationControlled2002}.
\item Robustness: Protection mechanisms to safeguard against the inconsistency or loss of essential data owing to system failures or network disruptions~\cite{lorenzHowFarFar2020, diasMarketBasedMultirobotCoordination2006, garciaArchitectureDecentralizedCollaborative2018}.
\item Reliability: Enable agnostic data sharing among robots to ensure efficient coordination~\cite{diasMarketBasedMultirobotCoordination2006}.
\item Openness: Foster a mission's public network environment, enabling relevant parties to participate on the platform~\cite{zlotMultirobotExplorationControlled2002}.
\item Usability: Accessible job request statuses and pricing information for processing and decision-making~\cite{zavolokina2018incentivizing}.
\item Compatibility: Ensure metadata requirements-driven data standards to diverse mapping tasks and applications~\cite{diasMarketBasedMultirobotCoordination2006}.
\item Maintainability: System design incentivizes participants to take ownership of maintaining the network~\cite{garciaArchitectureDecentralizedCollaborative2018}.
\item Portability: Keeping hardware requirements at a minimum to enable flexibility in deploying robot missions~\cite{rizkCooperativeHeterogeneousMultirobot2019, roehr2014reconfigurable, aditya2021survey}.
\item Interoperability: Ensure the system can handle traded data without significant adjustments~\cite{quintonMarketApproachesMultiRobot2023a, zlotMultirobotExplorationControlled2002, diasMarketBasedMultirobotCoordination2006}.
\end{enumerate}

\subsection{Architecture}

\begin{figure}
	\centering
	\includegraphics[width=0.99\columnwidth, clip, trim=110 30 105 10]{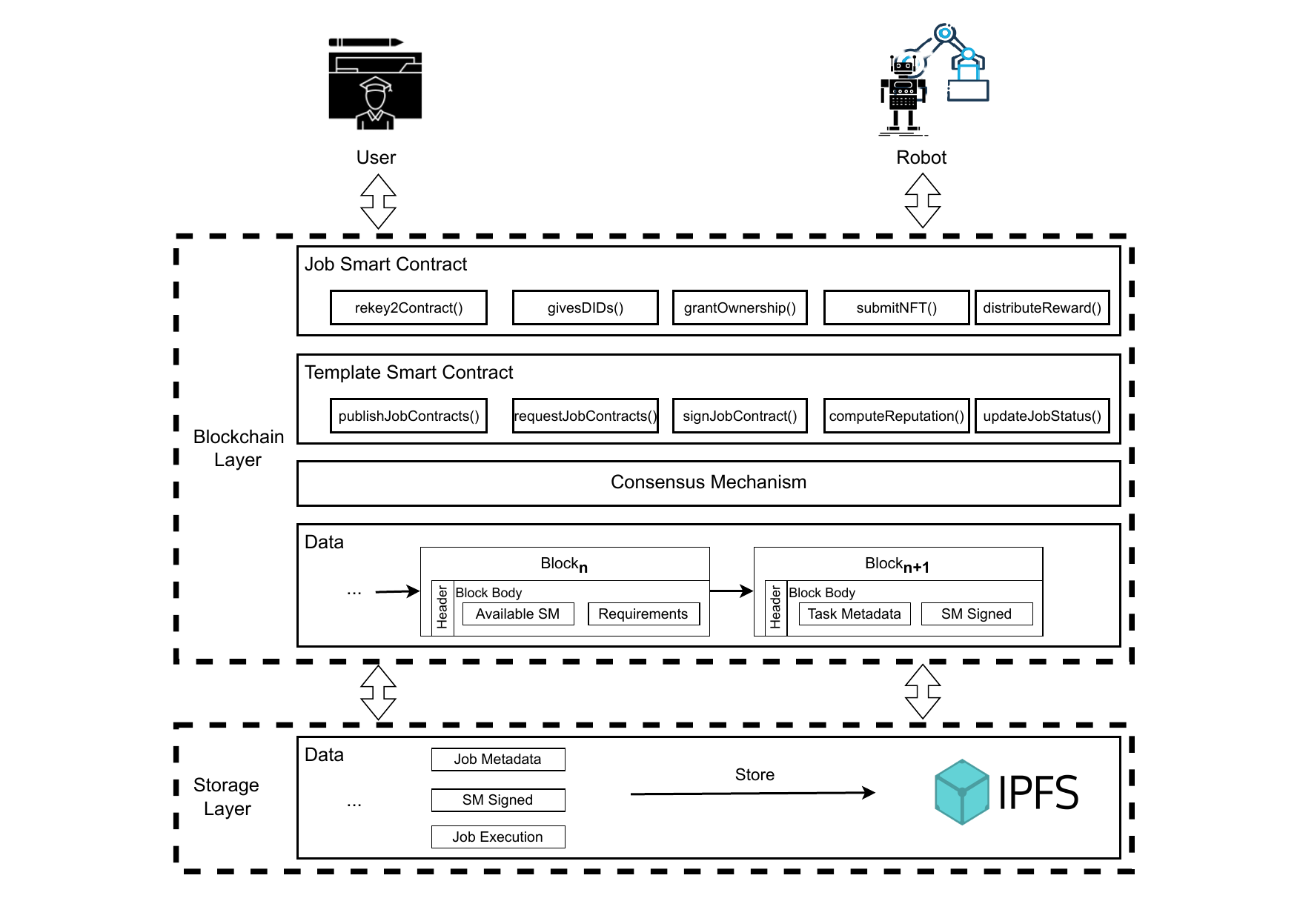}
	\caption{Architecture layers.}
     \label{fig:ArchitecturalLayers}
     \Description{A block diagram representing the architectural layer with three layers interacting with each other, the upper layer representing the user and robot, the second (middle) and third (bottom) layers are block diagrams representing the blockchain and the storage layer, each representing the components, composed of job contracts and smart contract template with each function, and the data layer components}
\end{figure}

Figure~\ref{fig:ArchitecturalLayers} features an overview of the architecture layers and their functions in our decentralized system architecture for job requests and contract creation. The process involves various entities, including the Client, MeshNetwork, Robot, TemplateSmartContract, JobSmartContract, and JobContract. The lifecycle begins with the Client initiating a JobRequest by sending a message to the MeshNetwork. The MeshNetwork broadcasts the request to available Robots, and each Robot optimizes and schedules its operation outside the blockchain. Once the scheduling is complete, the Robot sends the JobRequest through the TemplateSmartContract interface.
Upon receiving the JobRequest, the TemplateSmartContract creates a JobRequest's unique identifier. The TemplateSmartContract then shares the ID (transaction hash) of the JobRequest with the Robot. If at least one Robot decides to participate, a JobContract is generated and submitted to the TemplateSmartContract. The TemplateSmartContract updates the proposed job options from the JobContract with the most cost-effective proposal. The ID of the lowest-priced JobContract is returned and linked to the corresponding JobRequest, while it informs the current status and price to the Client.

\section{Evaluation}
\label{sec:evaluation}

This section has two subsections that examine the feasibility and performance of our proposed architecture. In subsection~\ref{sec:requirement_validation}, we focus on logically validating the architectural requirements of our design. We discuss network communication, data transparency, data integrity assurances, and the use of smart contracts for automated economic decision-making. In subsection~\ref{sec:Simulation}, we evaluate our system’s fitness to meet accuracy targets in mapping tasks and resilience in replicated celestial conditions through simulations and physical experiments. This subsection also discusses usability and transaction execution duration aspects.

\subsection{Requirements validation}
\label{sec:requirement_validation}

Our architecture, depicted in Figure~\ref{fig:ArchitecturalLayers}, integrates the network infrastructure with~\ac{DLT} for transparent, immutable, and \emph{coopetitive} robot coordination. By utilizing an \ac{IPFS}, the platform stores only image hashes that define the images' URLs on the blockchain, ensuring data integrity. To ensure conformity~($NFR-1$), the economic automated decision-making is based on the make-or-buy decision framework~\cite{canezDevelopingFrameworkMake2000}. The JobContract~($FR-4$), which records the execution of mapping tasks and facilitates information trading, and the TemplateSmartContract acting as an interface for robots to post and monitor JobRequests~($FR-5$), facilitating efficient robot coordination and automating job assignment and compensation~($FR-1$) in a non-discriminatory environment~($FR-7$).

We initially used a mesh network for communication and control systems suitable for decentralized peer-to-peer and low-latency settings~($NFR-3$). However, we identified exploitation vulnerabilities regarding unauthorized data modification during transmission between peers. Given the high knowledge-sharing capabilities of \emph{coopetitive} platforms~\cite{ghobadiCoopetitiveKnowledgeSharing2011b}, the system's integration allows for the secure and transparent sharing of mapping data among the participating robots, where robots add the data requirements (i.e., mapping location) to be validated by the smart contract~($NFR-4$) via the specific stratified Goldberg polyhedron registry position approach~\cite{gooDIANABlockchainLunar2019}. When followed, it ensures the reliability of the collected information~($FR-10$). The system publicly records who was the pioneer in being the first to explore, investigate, or discover an area or execute activities recognized as at the forefront of space exploration initiatives. As such, this verifiable record can help to build valuable partnerships and position universities as leaders in innovative interdisciplinary research. Additionally, a commission model can compensate the pioneer with every sale of the \ac{NFT} associated with the map on secondary markets~\cite{hartwichProbablySomethingMultilayer2023}, translating into additional revenue streams.

\ac{IPFS} enables robust and decentralized storage of mapping data ($NFR-2$) and protects from loss or censorship~($FR-8$). It accepts standard data formats~($NFR-2$) and detects tampering through hash verification. While transactions within the blockchain are immutable, standard asset parameters can be modified. With these parameters, entities can confidently re-configure, destroy, mint, freeze, and even clawback addresses, ensuring that only authorized participants can modify the mapping metadata within the system~($NFR-7$). The history of these assets is immutably recorded in the blockchain and can be publicly verified. In our case, these systems do not depend on specific specialized hardware and enable near real-time processing in space-located facilities or, as we have successfully implemented, in a server-client structure that simplifies data sharing~($NFR-8$). Thus, it affirms the supposed improved performance of a distributed, decentralized system consisting of multiple \emph{coopetitive} robots compared to single individual mapping units.

During the bidding process, our robots engage in a descending-price auction. By posing a lower bid than the previous lowest bid recorded in the smart contract~($FR-2$), robots signal their willingness to execute a job. This descending-price bidding allows the robots to continuously reassess their capabilities and resources, resulting in an economically efficient solution~\cite{diasMarketBasedMultirobotCoordination2006}. The robots can negotiate multiple contracts simultaneously, and different robots can initiate new contracts to find new ways to execute previous contract proposals~($NFR-5$). As a result, several JobRequests may attach to multiple JobContracts, coordinated by several TemplateSmartContract instantiations~($NFR-9$). The TemplateSmartContract evaluates the proposals and ranks the options as it reach the time limit. The Client then considers the JobContract and decides whether to accept it. If accepted, an atomic payment transfer process is initiated by rekeying the contract signature to the TemplateSmartContract, potentially triggering multiple intermediate contracts. The respective bidding robot can set a maximum price threshold to ensure that auctioning outcomes align with it.

Regarding exploitation protection in this phase, to ensure the contract is not vulnerable to skipping payment after task execution, it is only considered ready for execution once the TemplateSmartContract receives the rekey power. Once granted, the contract can be signed, and the payment can be made automatically according to the contract agreement. However, if an attacker keeps bidding lower than others, there is no automated protection against spamming or the need to create new auctions. The solutions hence incorporate the bidder’s reputation, which may be lower than others, and setting a time limit to restrict possible spamming attacks. These attacks may involve proposing multiple fake tasks or submitting fake lower bids, which the robot never intended to execute, or it just wanted to prevent the transaction from happening. Another potential exploitation is the failure to complete tasks upon which parties previously agreed. Even though our robots are automatically programmed to function independently, any deliberate harmful human interference can still disrupt their operations. Nevertheless, the physical limitations on communication time delays from Earth and the celestial body and the smart contract time limit may offer some protection against such cases. Despite unintended malfunctions that may still occur, this physical protection, combined with the reputations of bidders, can improve trust in the platform.

\begin{figure}[h]
	\centering
	\includegraphics[width=\columnwidth]{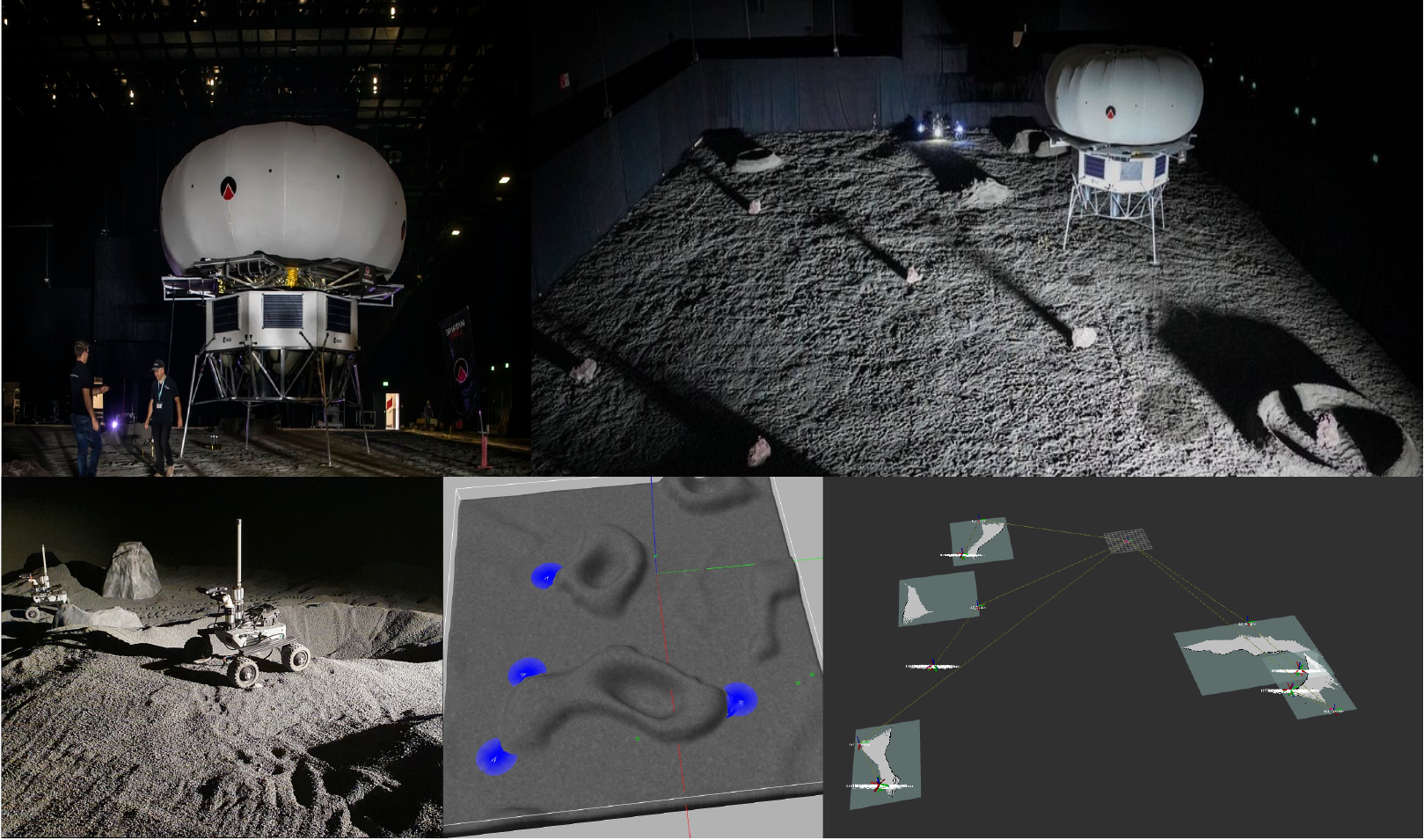}
	\caption{Experimental environment that mimics the environmental lunar conditions (Lunalab), the \Acl{EL3} and the \emph{Gazebo}'s~\cite{koenigDesignUseParadigms2004} virtual space for multi-robot simulation integrated with \emph{ROS~2}~\cite{quigleyROSOpensourceRobot2009}.}
    \label{fig:MoonLab}
     \Description{A collection of images featuring the lunar laboratory, which includes lunar rovers and the European cooperative large logistic lander. It also presents the virtual simulation environment and the mapping retraction software.}
\end{figure}

Once the payment is received, the TemplateSmartContract triggers the participating robots to sign and execute the JobRequest and JobContract~($FR-4$). The robots carry out the specified tasks outlined in the contract: mapping or selling the map exploration rights while adding the asset description into the metadata and data~($NFR-6$). Before finalizing the contracts, the TemplateSmartContract conducts a plausibility check of the correctness and completeness of the mapping data retrieved from \ac{IPFS} by assessing the requirements described on the JobRequest and the \ac{NFT} metadata. This check ensures that the data adheres to the initial specifications outlined in the JobRequest without any unauthorized modifications, such as via the mutable asset characteristics. The possible mutability functions are: freezing an asset until the user meets a specific requirement, clawback by debiting a user’s account for defaulting in loan payments, and revoking the ownership of assets belonging to users. Thus, the completion is successful if the robot receives the payment according to the due terms of the JobContract.

In the last stage, due to the \acp{NFT}' (meta-) data, the Client is recognized as the pioneer explorer of the acquired mapping data or deliverables, signifying the process completion. The TemplateSmartContract verifies that the retrieved map metadata and \ac{IPFS} data align with the original JobRequest and the acquired mapping information~($FR-3$). After the verification process, the Client can use the data for decision-making, knowing that the mapping coordination process was transparent and trustworthy. We implemented a prototype of our designed architecture to evaluate its practicality and discuss the lessons learned.

\subsection{Requirements verification in simulation and implementation}
\label{sec:Simulation}

The evaluation of our system aimed to test and validate its accuracy and resistance in similar celestial conditions, explicitly focusing on assessing space-related limitations, such as latency and resource constraints. We conducted three simulations to evaluate the architectural solution's ability to collect, coordinate, and maintain mapped area data. We started with a \texttt{turtlesim}-based simulation\footnote{\url{https://wiki.ros.org/turtlesim}} during the conceptualization stage using \emph{ROS~2}~\cite{quigleyROSOpensourceRobot2009}. While \emph{ROS~2} simplified the complex robotic development with open-source software tools and libraries, in the second simulation, we used \emph{Gazebo}’s realistic 3D~simulation platform~\cite{koenigDesignUseParadigms2004} for testing robot models and algorithms, thus providing a closer reproduction of the final simulated environment. Additionally, as virtual simulation may not sufficiently replicate real-world situations~\cite{garciaRoboticsSoftwareEngineering2020}, the third evaluation represented a proof-of-concept conducted in a celestial-like environment laboratory setting. Each test had an approximate duration of 30~minutes. The robotic platform utilized a visual-SLAM algorithm during the last assessment to create a map of the analogous lunar terrain facility, as illustrated in Figure~\ref{fig:MoonLab}. The facility in which we trained the robots to create precise maps, an $80\,\mathrm{m}^2$ rectangle filled with basalt, was designed to emulate the surface of the Moon's south pole, an area where researchers expect to find resources.

\begin{figure*}
	\centering
    \includegraphics[width=\textwidth]{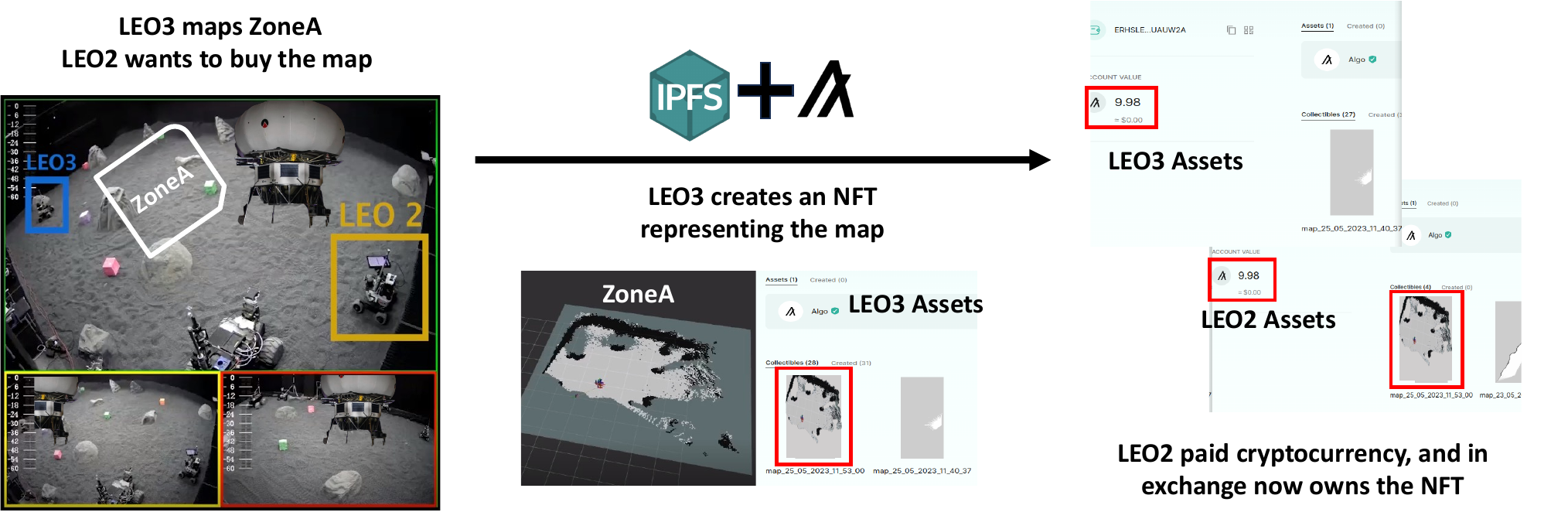}
	\caption{Prototype in the laboratory of simulated Moon environment.}
    \Description{A collection of images featuring the architectural sequence implementation. The first image represents the lunar laboratory described in Figure~\ref{fig:MoonLab}, with the LEO3 mapping ZoneA and LEO2 wanting to buy the map. This represents the scenario of robots as depicted in Figure~\ref{fig:Moonmap}. It follows the sequence with LEO3, creating the NFT that represents the map of ZoneA. Finally, the last two images on the left portray LEO2 and LEO3 assets wallets, representing that LEO2 pays the cryptocurrency to LEO3, and in exchange, LEO2 now owns the NFT that LEO3 transferred.}
    \label{fig:Prototype}
\end{figure*}

As we are currently in the prototype phase, our primary focus is to assess the technical feasibility and performance of the architecture. Our system's architecture, illustrated in Figure~\ref{fig:Prototype}, is designed for efficient data storage, \ac{NFT} management, and trading with 3~stages: \ac{NFT} creation, sale, and retrieval. We harnessed the capabilities of \emph{Python}, $C++$, \emph{Gazebo}, and \emph{ROS~2} for development, ensuring effective communication and simulation. To bridge the communication gap between robots (LEO2 and LEO3) and the blockchain, we integrated the PureStake connector via a \emph{REST} service~\footnote{\url{https://www.purestake.com/}}. 

All communication flows, except for the smart contract, have been implemented, encompassing the autonomous creation, negotiation, coordination, selling, and verification of \acp{NFT} and mapping requirements. Despite not taking full advantage of the smart contract automated behavior, the prototype already leverages a blockchain network for trading \acp{NFT} and certain data storage aspects, enabling real-time decision-making based on blockchain data. The architecture uses \ac{IPFS} for high-throughput content-addressed block storage~\cite{benetIPFSContentAddressed2014}. Entities can govern their mapping \acp{NFT} metadata and settings, from transferability up to their destruction, promoting value to a market for data access and recognizing pioneering exploration. These \acp{NFT} encapsulate specific map data and metadata, containing coordinates, resolution, sensors, mapping algorithms, and map price, fostering trade within the network and improving \ac{ISRU}'s information symmetry.

Our evaluation assessed the system's bandwidth usage and transaction duration in simulated lunar network conditions. Despite transactions needing to be fully optimized, the average bandwidth usage during uploads was less than 5\,MB over 4~seconds, with a median of 4.7\,MB. The standard deviation was 0.5\,MB, indicating a relatively consistent transaction performance. The time required for uploading maps and conducting sales transactions was approximately 10~seconds, with a median of 9~seconds. The standard deviation was 2~seconds, indicating that most transactions were completed within a reasonable time frame. These durations are relatively short compared to the time robots typically take to navigate to the place and process cartographic data, which, at best, can take several minutes to hours. As a result, the upload duration, plus mapping and reaching a consensus, is negligible compared to considering the inherent latencies of bidirectional communication between terrestrial stations and robotic units (which, for the Moon case, takes a minimum of five seconds as the signal needs to pass through the different relays on Earth and satellites in space), let alone the possibility of other nations being open to start cooperating. 
The system displayed reasonable transaction duration usage and significantly reduced robot coordination delays compared to waiting for human coordination.

\section{Discussion}
\label{sec:discussion}

Our evaluation demonstrates the technical feasibility of the proposed architecture for \ac{MRS} in collaborative scientific space exploration. The system successfully collects, coordinates, and maintains mapped area data via the blockchain metadata while evaluating the mean bandwidth usage and transaction duration to validate the feasibility in real-world testing conditions (see Figure~\ref{fig:Prototype}). Moreover, our architecture 
offers precise map data that can be leveraged for various applications, including celestial terrain coverage with sustainable \ac{ISRU} capabilities, thereby enhancing information symmetry and decision-making processes. %

As for scalability, while the initial setup requires resources to bootstrap the constellation map through multiple interactions, the system is designed to handle the growing data in \ac{DLT} or \ac{IPFS} without being overly strained. The blockchain only stores limited map metadata, managing rights, properties, and descriptions. Nevertheless, there may be challenges with multiple mapping data streams in the \ac{IPFS} system in the long run. 

\subsection{Limitations}
\label{sec:limitations}

This section outlines the limitations of our architecture, categorized into technical challenges, economic implications, and future research directions.
Ensuring the reliability and integrity of shared mapping data and the effectiveness of smart contracts remains a challenge. As such, future work should focus on measurement systems, quality assurance metrics, and addressing the practical limitations of smart contracts in decentralized systems~\cite{afanasyevBlockchainSolutionsMultiAgent2019}. For instance, validation mechanisms~\cite{liBlockchainBasedCrowdsourcingFramework2022}, reputation systems~\cite{calvaresiTrustedRegistrationNegotiation2018, yuSurveyMultiagentTrust2013}, and different consensus mechanisms~\cite{castelloferrerBlockchainNewFramework2019, rizkDecisionMakingMultiagent2018} could be integrated to enhance mapped data reliability. Continuous monitoring and anomaly detection algorithms can also support detecting and mitigating suspicious activities~\cite{strobelManagingByzantineRobots2018}. %
It is crucial to understand the practical implications of public vs. private networks, permissionless vs. permissioned designs, and the specific consensus mechanisms that minimize computational burden, specifically for resource-constrained space \ac{MRS}~\cite{Alfraheed_Al-Zaghameem_2013}.

Although our implementation aims to solve the dependable communication limitation for \ac{DLT}-based platforms to thrive, the automated market-based approaches for \ac{MRS}, such as blockchain auctioning systems, are still in the early stages. Auction systems are also considered intricate and less effective in centralized situations and excessively complex in distributed situations~\cite{diasMarketBasedMultirobotCoordination2006}. Additionally, hierarchical decision-making and external third-party systems can pose challenges in identifying trustworthy evaluators for smart contract task execution~\cite{liBlockchainBasedCrowdsourcingFramework2022, ryosukeabeFabchainManagingAuditable2022}. Further research should investigate efficient communication protocols for blockchain, smart contracts, market mechanisms~\cite{diasMarketBasedMultirobotCoordination2006}, and decentralized combinatorial optimization systems~\cite{baldorApplyingCloudComputing2013}, which are crucial for space missions.

The economic implications of \ac{DLT} in space \ac{MRS} require further investigation and support from real-world examples and case studies. 
Narratives for blockchain technology are essential for promoting its acceptance of new ways of governing outer space~\cite{filippiBlockchainOuterSpace2021}. Real-world examples and case studies exploring economic interactions facilitated by \ac{DLT} in space exploration, such as resource trading, service provisioning, and fair compensation mechanisms, would enhance the credibility and depth of the discussion~\cite{baldesiBitcoinLeveragingBlockchain2017, filippiBlockchainOuterSpace2021, ibrahimLiteratureReviewBlockchain2021}. Future research should focus on real-world deployments and economic modeling to quantify the economic impact of our architecture.
 
As~\citet{afanasyevBlockchainSolutionsMultiAgent2019} point out, the field must transition from theoretical discussions to practical implementations. Our work aims to contribute to this transition by implementing prototypes in celestial-like conditions. We are addressing the limitations and challenges to ensure the platform's efficiency and economic viability in space exploration. We intend to continue our research and development efforts to address these challenges, keeping in line with the increasing interest from the scientific community, space agencies~\cite{baldesiBitcoinLeveragingBlockchain2017, israelSpaceGovernance2020}, and industry~\cite{ibrahimLiteratureReviewBlockchain2021}.

\subsection{Research gap and open challenges}

\textbf{High autonomy in limited resources:} \emph{Coopetitive} platforms are known to provide high knowledge-sharing capabilities~\cite{ghobadiCoopetitiveKnowledgeSharing2011b}, where sophisticated techniques, such as ontologies, semantic web, and linked data~\cite{memduhogluPossibleContributionsSpatial2018} would be an option, given that robots can learn new abilities with minimal incremental rewarded data~\cite{baima_2021_modeling}. However, it is challenging in environments with limited resources to deeply understand data and manage its heterogeneity while ensuring system stability and safety.

\textbf{Global cooperation and open networks:} A neutral \emph{coopetitive} platform's economic aspect may also benefit companies by allowing their robots to perform low-priority tasks during idle time, potentially increasing equipment utilization and return on investment and improving global \ac{ISRU} efficiency. Collaboration is crucial in reducing mission costs and achieving economies of scale~\cite{chowHowFallingLaunch2022}. Despite early proposals to utilize planning knowledge to improve performance and efficiency~\cite{muppasaniPlanningOntologyRepresent2023a} and possibly learn incremental actions~\cite{baima_2021_modeling}, it is crucial to establish ways to exchange norms, maintain team identity, and promote group member confidence~\cite{strobelManagingByzantineRobots2018}.

\textbf{Reliable communication mechanisms and automated \ac{DLT}-based markets:} Hierarchical distributed decision-making structures have their strengths and weaknesses for \ac{DLT}-based automated markets~\cite{yan2013survey} 
but relying on external third parties still challenges identifying trustworthy evaluator methods for \ac{DLT}-based application layers, which could be an organization or a digitally supplied system challenges to identify trustworthy evaluator methods for \ac{DLT}-based application layers as it may reduce the trustworthiness of the network.

\textbf{Economic and legal aspects remain open:} While our research primarily focuses on the technical feasibility and implementation of blockchain technology in \ac{MRS} for space exploration, we acknowledge the significance of economic and legal aspects in this domain. The economic implications, including cost-benefit analysis, funding models, and the financial viability of such systems, remain crucial areas for future exploration. Likewise, legal considerations, encompassing regulatory compliance, space law, and data governance, pose complex challenges that are beyond the scope of this paper. 

\section{Conclusions}
\label{sec:conclusion}

Our primary scientific contribution lies in the architecture and requirement engineering for blockchain applications in \ac{MRS}. The prototype, while a crucial element for demonstrating the feasibility of our approach, is secondary to the theoretical framework we have adopted.
Our architecture, developed through interdisciplinary perspectives from information systems, engineering, and economics, answered the \textbf{RQ} by demonstrating the potential of \ac{DLT}. Our platform's flexibility and non-discriminatory aspect allow for leveraging idle resources through a robot-as-a-service platform. The prototype enables efficient mapping coordination via smart contracts, ensures data integrity via \acp{NFT}, and facilitates revenue generation through market-based cryptocurrency and \ac{NFT} transactions, allowing organizations to trade valuable assets in space missions.
Blockchain is employed not as an end in itself but as a tool to facilitate secure and reliable information exchange and coordination among robots, addressing key challenges in robotic communication and operation
Our research provides a foundation for future advancements and implementations, following the principles and guidelines of the \ac{DSR} approach. 

While our research presents promising results, several areas for further exploration and improvement remain. The economic implications of \ac{DLT} in space \ac{MRS} require more profound analysis, including identifying suitable use cases, evaluating economic incentives, and examining governance and legal considerations. Additionally, while addressing the environmental impact of \ac{DLT} in space missions, future research should evaluate the performance and efficiency of our approach, consider transaction throughput and latency, optimize energy efficiency, and explore scalability solutions such as sharding and layer two protocols~\cite{hartwichMachineEconomies2023}. Future research could also develop decision support on when \ac{DLT} is required instead of digital signatures and bilateral communication to securely and efficiently implement concurrent access. However, this dilemma again emphasizes that while a \ac{DLT} may be desirable from an economic perspective, it can add substantial complexity to design and setup. Thus, it is crucial to identify suitable use cases for \ac{DLT} in space \ac{MRS} and understand service providers' economic needs and incentives.

\begin{acks}
This research was funded in part by the \grantsponsor{GS501100001866}{Luxembourg National Research Fund (FNR)}{https://doi.org/10.13039/501100001866} in the FiReSpARX Project, ref. \grantnum{GS501100001866}{14783405}, PABLO Project, ref. \grantnum{GS501100001866}{16326754}, and by PayPal, PEARL grant reference \grantnum{FNR}{13342933}/ Gilbert Fridgen. For the purpose of open access, and in fulfillment of the obligations arising from the grant agreement, the author has applied a Creative Commons Attribution 4.0 International (CC BY 4.0) license to any Author Accepted Manuscript version arising from this submission.

\end{acks}

\bibliographystyle{ACM-Reference-Format}
\bibliography{bibliography} 

\end{document}